\documentclass{amsart}
\usepackage{amssymb}
\usepackage{amsthm}
\usepackage{amsmath}
\usepackage{amsfonts}

\usepackage{placeins}
\usepackage{url}
\usepackage{enumerate}
\usepackage{physics}
\usepackage[titletoc]{appendix}
\usepackage{fancyhdr}
\usepackage{xcolor}
\usepackage{comment}
\usepackage{tikz}
\usepackage{pgfplots}
\usepackage{tikz-cd}
\usepackage{float}
\usepackage{hhline}
\usepackage{array}
\usepackage[margin = 1in]{geometry}
\usepackage[round,comma,authoryear]{natbib}
\let\cite\citep
\usepackage[hidelinks]{hyperref}
\usepackage{setspace}
\usepackage{multirow}
\usepackage{adjustbox}
\usepackage{rotating}
\usepackage{amsmath}
\usepackage{framed}
\usepackage{graphicx}
\usepackage{subcaption}
\usepackage{adjustbox} 
\pgfplotsset{compat=1.17}
\pgfplotsset{colormap={violet}{rgb255=(25,25,122) rgb255=(238,140,238) color=(white)}}
\spacing{1.2}

\allowdisplaybreaks
\numberwithin{equation}{section}
\newtheorem{definition}{Definition}[section]

\newtheorem{result}[definition]{Result}

\newtheorem{remark}[definition]{Remark}
\newtheorem{desc}[definition]{Description}

\begin{document}
\title[Description of the Training Process of Neural Networks via Ergodic Theorem : Ghost nodes]{Description of the Training Process of Neural Networks via Ergodic Theorem : Ghost nodes}
\author[E.Park]{Eun-Ji Park}
\address{Graduate School of Data Science, Seoul National University, Gwanak-ro 1, Gwanak-gu, Seoul 08826, Korea}
\email{dmswl4743@snu.ac.kr}

\author[S.Yun]{Sangwon Yun}
\address{Department of Mathematical Sciences, Seoul National University, Gwanak-ro 1, Gwanak-gu, Seoul 08826, Korea}
\email{ysw317@snu.ac.kr}

\begin{abstract}
Recent studies have proposed interpreting the training process from an ergodic perspective. Building on this foundation, we present a unified framework for understanding and accelerating the training of deep neural networks via stochastic gradient descent (SGD). By analyzing the geometric landscape of the objective function we introduce a practical diagnostic, the running estimate of the largest Lyapunov exponent, which provably distinguishes genuine convergence toward stable minimizers from mere statistical stabilization near saddle points. We then propose a ghost category extension for standard classifiers that adds auxiliary ghost output nodes so the model gains extra descent directions that open a lateral corridor around narrow loss barriers and enable the optimizer to bypass poor basins during the early training phase. We show that this extension strictly reduces the approximation error and that after sufficient convergence the ghost dimensions collapse so that the extended model coincides with the original one and there exists a path in the enlarged parameter space along which the total loss does not increase. Taken together, these results provide a principled architecture level intervention that accelerates early stage trainability while preserving asymptotic behavior and simultaneously serves as an architecture-friendly regularizer.
\end{abstract}

\maketitle

\section{Introduction}

The training of modern neural networks is usually described through the lens of optimization, yet in practice the learning dynamics are driven by stochastic, high-dimensional iterations whose long-term statistical behavior is still only partially understood.  Recent work has shown that viewing stochastic gradient descent (SGD) as a Markov process on a compact weight space allows one to invoke classical ergodic theorems and obtain invariant measures that describe the distribution of network parameters over time. In this paper we adopt that ergodic-theoretic perspective from \citet{ZLSJ22} and \citet{VGCX24}, introducing a concrete, computable diagnostic inspired from \citet{Zhang23}: the running estimate of the largest Lyapunov exponent
\begin{align*}
\hat{\gamma}_{N} = {\frac{1}{N}}\sum_{t=0}^{N-1}\log\,\lVert I-\eta\nabla^{2}f({w}^{(t)})\rVert,
\end{align*}
whose sign reliably distinguishes genuine convergence toward stable minimizers from mere statistical stabilization around saddle regions.  In addition, we propose ghost category extension of standard classifiers, showing that the extra degrees of freedom create escape directions that accelerate early training without altering the eventual invariant law. Ghost extension provides the following: it initially employs large steps to bypass barriers via a ghost-coordinate shortcut toward a local minimizer; as ghost nodes are removed, the expanded space contracts to reinstate the original loss surface; and through ergodic averaging over many iterations, the average loss realigns with the original function, achieving convergence in fewer steps. Taken together, these results offer a unified probabilistic description of learning dynamics and open a path toward principled architecture-level interventions that improve training speed while preserving asymptotic behavior.

As a new approach to practical training improvements, we attach ghost nodes to control the optimization path through structural expansion of the output layer. Unlike classic regularizations, this architectural intervention naturally serves as a model-friendly regularizer: the ghost nodes fulfill their role and then dissipate over time, providing seamless regularization without additional ad hoc mechanisms.

\vskip 4mm

\section{Stochastic Gradient Descent Function}

In the problem of minimizing an objective function $f: \mathbb{R}^d \rightarrow \mathbb{R}$, our goal is to find the optimal solution $w^* = \arg\min_{w \in \mathbb{R}^d} f(w)$. However, computing the gradient $\nabla f(w)$ exactly over the entire dataset at each iteration is computationally expensive, particularly for large-scale problems. To address this issue, we consider the stochastic approach which follows from \citet{ZLSJ22}.

Instead of the deterministic gradient descent, we will mathematically review the following SGD (for more detailed content, see \citet{ZLSJ22}). At each iteration, we randomly select a mini-batch of size $m$ from the dataset, indexed by $\{i_1, i_2, \dots, i_m\}$, and compute the gradient estimator based on this mini-batch:
\[
g(w,\xi) = \frac{1}{m}\sum_{j=1}^{m}\nabla \ell(w;x_{i_j}),
\]
where $\ell(w;x)$ denotes the per-sample target function and $\xi$ represents the randomness arising from the mini-batch selection. That is, the function $\ell(w; x)$ denotes the loss on an individual sample $x$, whereas the objective function $f(w)$ is defined as the empirical average of $\ell$ across all samples in the dataset. By construction, this stochastic gradient estimator satisfies the unbiasedness condition $\mathbb{E}_\xi[g(w,\xi)] = \nabla f(w)$.

Given a fixed learning rate $\eta > 0$, the stochastic gradient descent (SGD) process starts from an initial point $w^{(0)}$ and iteratively updates as follows:
\[
{w}^{(n+1)} = {w}^{(n)} - \eta\,g({w}^{(n)},{\xi}_n),
\]
where each $\xi_n$ is independently drawn from the same distribution at every iteration. 

We now consider the following setting to analyze this stochastic gradient descent procedure from the perspective of ergodic theory. We make the assumption that $W \subseteq \mathbb{R}^d$ is a compact parameter space large enough that all iterates remain inside, and that the objective function $f: W \rightarrow \mathbb{R}$ is continuously differentiable. Under this setting, we define the stochastic gradient descent map $D : W\times\Xi \rightarrow W$ as
\begin{align}\label{SGD map}
D(w,\xi) := w - \eta\,g(w,\xi),
\end{align}
for each $w \in W$ and random variable $\xi$ representing the mini-batch selection.

In order to apply the ergodic theorem, we require two important conditions: (i) the stochastic map $D$ maps the parameter space $W$ into itself, i.e., $D(W,\xi)\subseteq W$ almost surely, and (ii) the mapping $(w,\xi)\mapsto D(w,\xi)$ is continuous in $w$ for each fixed $\xi$. Condition (ii) follows directly from the continuity and differentiability of the loss function $f$. Regarding condition (i), we can easily ensure this by choosing a sufficiently small learning rate $\eta$ and, if necessary, slightly extending $W$ beyond its natural boundary. Specifically, we extend the parameter space $W$ slightly and modify the loss function $f$ to increase sharply near the boundary of the extended region, ensuring differentiability and maintaining the iterates within the original domain $W$.

By constructing the stochastic gradient descent map in this way, we establish a well-defined Markov process. Consequently, standard ergodic theory arguments for Markov chains guarantee the existence of an invariant probability measure supported on $W$.

\vskip 4mm

\section{Basic Preliminaries of Ergodic Theory}

In this section, we briefly introduce the basic preliminaries for the ergodic theorem to be used later, following the exposition in \citet{VGCX24}. Consider a compact parameter space $W\subset\mathbb{R}^d$, chosen large enough so that all iterates of the stochastic gradient descent remain inside. Fix a stepsize $\eta>0$ and let $(\xi_n)_{n\ge0}$ be an i.i.d.\ sequence of random variables whose distribution has a density that is strictly positive on a neighborhood of the origin. We considet the SGD map (see \eqref{SGD map}):
\[
  D\colon W\times\Xi\;\to\;W,\qquad
  D(w,\xi)=w-\eta\,g(w,\xi),
\]
where $g(w,\xi)$ satisfies $\mathbb{E}_\xi[g(w,\xi)]=\nabla f(w)$. Starting from $w_0\in W$, the iteration
\[
  {w}^{(n+1)}=D({w}^{(n)},{\xi}_{n})\quad(n=0,1,2,\dots)
\]
defines a time-homogeneous Markov chain on $W$ with transition probability
\[
  P(w,A)=\Pr\bigl(D(w,\xi_0)\in A\bigr)
\]
for any Borel subset $A\subset W$.

Moreover, the chain is ergodic, meaning its long-term behavior is independent of the starting point, and it satisfies the hypotheses of the Markov-chain pointwise ergodic theorem (Douc et al., 2018). Consequently, for any bounded continuous function $h\colon W\to\mathbb{R}$ we have almost surely
\[
  \lim_{N\to\infty}\frac{1}{N}\sum_{n=0}^{N-1}h({w}^{(n)})
  =\int_W h(w)\,\mu(dw).
\]
In particular, taking $h(w)=f(w)$ and $h(w)=\|\nabla f(w)\|$ yields
\[
  \frac{1}{N}\sum_{n=0}^{N-1}f({w}^{(n)})
  \;\xrightarrow{\mathrm{a.s.}}\;
  \int_W f\,d\mu,
  \quad
  \frac{1}{N}\sum_{n=0}^{N-1}\|\nabla f({w}^{(n)})\|
  \;\xrightarrow{\mathrm{a.s.}}\;
  \int_W\|\nabla f\|\,d\mu.
\]
Furthermore, the empirical distribution obtained by placing a unit mass at each iterate and averaging,
\[
  \frac{1}{N}\sum_{n=0}^{N-1}\delta_{{w}^{(n)}},
\]
converges weakly to $\mu$ almost surely (see equation (5) of \citet{ZLSJ22}).

\vskip 4mm

\section{Application in the learning process of neural networks}

In this section, we aim to develop practical content about the Lyapunov exponent, also based on the theoretical results of \citet{VGCX24}.

\begin{result}\rm
A practical diagnostic is to track the largest Lyapunov exponent
\begin{align}
  \gamma=\lim_{N \to \infty}\frac{1}{N}
         \sum_{n=0}^{N-1}\,
         \log\,\bigl\|I-\eta\nabla^{2}f({w}^{(n)})\bigr\|,
\end{align}
defined whenever
\(\mathbb{E}_{\mu}\bigl[\log\|I-\eta\nabla^{2}f\|\bigr]<\infty\) (see also \citet{Zhang23}).
\end{result}

\noindent

Especially, the running estimate
$\widehat{\gamma}_{N} = \frac{1}{N}\sum_{n=0}^{N-1}\log\,\|I-\eta\nabla^{2}f({w}^{(n)})\|$
can be computed and practical meaning that persistently
negative values suggest convergence, while values near zero or positive
show that the optimizer is still roaming. When \(f\) is strictly convex every Hessian eigenvalue is positive,
\(I-\eta\nabla^{2}f\) has spectrum inside the unit disk and \(\gamma<0\), that is, classical contraction toward the unique minimizer is recovered. The Lyapunov-based indicator then supplies a concrete test for distinguishing genuine convergence from mere statistical stabilization during training.

From the above practical test that distinguishes convergence to genuine minimizers from stagnation at saddle points, thereby capturing geometric information about the target landscape that extends beyond purely local structure. In doing so, we leverage local information from the target function while aiming to adjust the restricted region of the parameter space to capture geometric information at a larger scale beyond the mere local structure.

\vskip 4mm

\section{Description of Ghost Nodes via Ergodic Theorem}\label{sec:ghost}

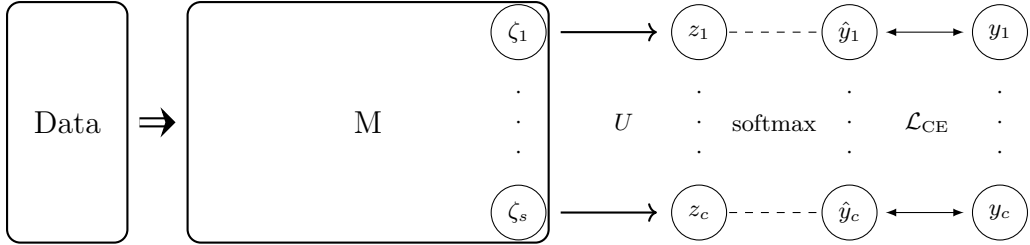
\begin{figure}[h]
\centering
\begin{tikzpicture}[font=\small, scale = 0.8, circl/.style={circle, fill = black, inner sep = 2pt}]
\draw[rounded corners=5, thick] (1,0) rectangle (3,4);
\node[draw=none, font = \Large] at (2, 2) {$\textup{Data}$};
\draw[rounded corners=5, thick] (4,0) rectangle (10,4);
\node[ draw=none, font = \Large] at (7, 2) {$\textup{M}$};
\node[circle={3pt},draw = black] at (9.5,3.5){$\zeta_1$};
\node[draw=none] at (9.5,2.5){$\cdot$};
\node[draw=none] at (9.5,2){$\cdot$};
\node[draw=none] at (9.5,1.5){$\cdot$};
\node[circle={2pt},draw = black] at (9.5,0.5){${\zeta}_{s}$};
\draw[dashed] (13,3.5) -- (14.5,3.5);
\draw[dashed] (13,0.5) -- (14.5,0.5);
\node[draw=none] at (12.5,3.5){$z_{1}$};
\draw (12.5,3.5) circle (13pt);
\node[draw=none] at (12.5,2.5){$\cdot$};
\node[draw=none] at (12.5,2){$\cdot$};
\node[draw=none] at (12.5,1.5){$\cdot$};
\node[draw = none] at (12.5,0.5){$z_{c}$};
\draw (12.5,0.5) circle (13pt);
\node[draw=none] at (15,3.5){${\hat{y}}_{1}$};
\draw (15,3.5) circle (13pt);
\node[draw=none] at (15,2.5){$\cdot$};
\node[draw=none] at (15,2){$\cdot$};
\node[draw=none] at (15,1.5){$\cdot$};
\node[draw = none] at (15,0.5){${\hat{y}}_{c}$};
\draw (15,0.5) circle (13pt);
\node[draw=none] at (17.5,3.5){${y}_{1}$};
\draw[latex-latex] (15.6,3.5) -- (16.9,3.5);
\draw[latex-latex] (15.6,0.5) -- (16.9,0.5);
\node[draw=none] at (17.5,2.5){$\cdot$};
\node[draw=none] at (17.5,2){$\cdot$};
\node[draw=none] at (17.5,1.5){$\cdot$};
\node[draw = none] at (17.5,0.5){${y}_{c}$};
\draw (17.5,0.5) circle (13pt);
\node[draw = none] at (16.29,2){$\mathcal{L}_{\textup{CE}}$};
\node[draw = none] at (13.75,2){\textup{softmax}};
\node[draw = none] at (11.25,2){\textup{$U$}};
\draw (17.5,3.5) circle (13pt);
\draw[double,->, -{Stealth[length=2mm, width=4mm]}, thick,double distance=2.25pt] (3.2,2) -- (3.8,2);
\path[->, thick] (10.2,3.5) edge[] (11.8,3.5);
\path[->, thick] (10.2,0.5) edge[] (11.8,0.5);
\end{tikzpicture}
\caption{Description of the model $\textup{M} \otimes \textup{S}$}\label{fig: Diagram 1}
\end{figure}

We will now focus specifically on scenarios dealing with multi-class classification problem. Let $\textup{M}$ be a model based on the deep neural network. In this case, we can consider a model $\textup{M} \otimes \textup{S}$ that classifies the categories of a given data. In this context, $\textup{S}$ is a classifier that takes the last outputs of $\textup{M}$, denoted as $\zeta_{i}$, as its input. It applies the weights $u$ and an activation function $\sigma$, and then advances to a single hidden layer with softmax (see, Figure \ref{fig: Diagram 1}).

\vskip 2mm

More precisely, we can write the components of $\textup{M} \otimes \textup{S}$ as follows:
\begin{align}\label{eq: zeta and z}
{\zeta}_{j} = {\textup{M}}_{j},
{\quad}
{z}_{i} = {\sigma}({u}_{1,i}{\zeta}_{i} + \cdots + {u}_{s,i}{\zeta}_{s} + {b}_{i})
\end{align}
where $j$ (resp. $i$) ranges from $1$ to $t$ (resp. $c$), and $\mathcal{L}_{\textup{CE}}$ is a categorical cross entropy function. Then we can now consider a larger model $\textup{M} \otimes (\textup{S} \oplus {\textup{S}}^{gh})$, which includes the original model $\textup{M} \otimes \textup{S}$, as in the following description.

\begin{desc}[Extended categorical model $\textup{M} \otimes (\textup{S} \oplus {\textup{S}}^{gh})$]\label{desc: Main model}
\rm
Let $\textup{M} \otimes \textup{S}$ be the model given in Figure \ref{fig: Diagram 1}. Then we can construct the model $\textup{M} \otimes (\textup{S} \oplus \textup{S}^{gh})$, which we will call an extended categorical model, by considering the additional ghost categories in $\textup{S}^{gh}$. The submodel $\textup{S}^{gh}$ assigns an $e$ additional categories as illustrated in the diagram below:

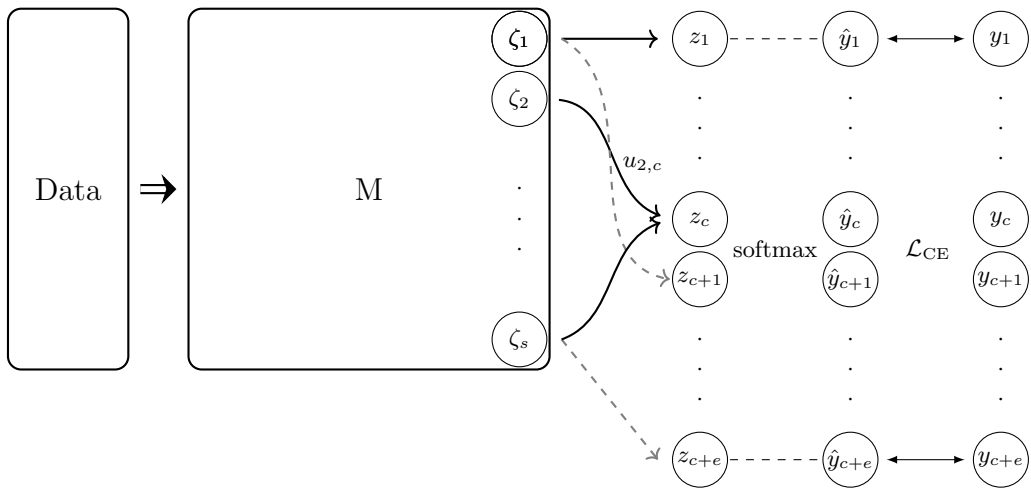
\begin{figure}[h]
\centering
\begin{tikzpicture}[font=\small, scale = 0.8, circl/.style={circle, fill = black, inner sep = 2pt}]
\draw[rounded corners=5, thick] (1,0) rectangle (3,6);
\node[draw=none, font = \Large] at (2, 3) {$\textup{Data}$};
\draw[rounded corners=5, thick] (4,0) rectangle (10,6);
\node[ draw=none, font = \Large] at (7, 3) {$\textup{M}$};

\node[circle={3pt},draw = black] at (9.5,5.5){$\zeta_1$};
\node[circle={3pt},draw = black] at (9.5,4.5){$\zeta_2$};
\node[draw=none] at (9.5,3){$\cdot$};
\node[draw=none] at (9.5,2.5){$\cdot$};
\node[draw=none] at (9.5,2){$\cdot$};
\node[circle={2pt},draw = black] at (9.5,0.5){${\zeta}_{s}$};
\draw[dashed] (13,5.5) -- (14.5,5.5);
\node[draw=none] at (12.5,5.5){$z_{1}$};
\draw (12.5,5.5) circle (13pt);
\node[draw=none] at (12.5,4.5){$\cdot$};
\node[draw=none] at (12.5,4){$\cdot$};
\node[draw=none] at (12.5,3.5){$\cdot$};
\node[draw = none] at (12.5,2.5){$z_{c}$};
\draw (12.5,2.5) circle (13pt);
\node[draw = none] at (12.5,1.5){{$z_{c+1}$}};
\draw (12.5,1.5) circle (13pt);
\node[draw=none] at (12.5,0.5){$\cdot$};
\node[draw=none] at (12.5,0){$\cdot$};
\node[draw=none] at (12.5,-0.5){$\cdot$};
\node[draw = none] at (12.5,-1.5){$z_{c+e}$};
\draw (12.5,-1.5) circle (13pt);
\draw[dashed] (13,-1.5) -- (14.5,-1.5);
\node[draw=none] at (15,5.5){${\hat{y}}_{1}$};
\draw (15,5.5) circle (13pt);
\node[draw=none] at (15,4.5){$\cdot$};
\node[draw=none] at (15,4){$\cdot$};
\node[draw=none] at (15,3.5){$\cdot$};
\node[draw = none] at (15,2.5){${\hat{y}}_{c}$};
\draw (15,2.5) circle (13pt);
\node[draw = none] at (15,1.5){${\hat{y}}_{c+1}$};
\draw (15,1.5) circle (13pt);
\node[draw=none] at (15,0.5){$\cdot$};
\node[draw=none] at (15,0){$\cdot$};
\node[draw=none] at (15,-0.5){$\cdot$};
\node[draw = none] at (15,-1.5){{${\hat{y}}_{c+e}$}};
\draw (15,-1.5) circle (13pt);

\node[draw=none] at (17.5,5.5){${y}_{1}$};
\draw[latex-latex] (15.6,5.5) -- (16.9,5.5);
\node[draw=none] at (17.5,4.5){$\cdot$};
\node[draw=none] at (17.5,4){$\cdot$};
\node[draw=none] at (17.5,3.5){$\cdot$};
\node[draw = none] at (17.5,2.5){${y}_{c}$};
\draw (17.5,2.5) circle (13pt);

\node[draw = none] at (16.29 ,2.0){$\mathcal{L}_{\textup{CE}}$};
\node[draw = none] at (13.75, 2.0){\textup{softmax}};
\node[draw = none] at (17.5,1.5){${y}_{c+1}$};
\draw (17.5,1.5) circle (13pt);
\node[draw=none] at (17.5,0.5){$\cdot$};
\node[draw=none] at (17.5,0){$\cdot$};
\node[draw=none] at (17.5,-0.5){$\cdot$};
\node[draw = none] at (17.5,-1.5){{${y}_{c+e}$}};
\draw[latex-latex] (15.6,-1.5) -- (16.9,-1.5);
\draw (17.5,-1.5) circle (13pt);
\draw (17.5,5.5) circle (13pt);
\node (ZE2) at (10, 4.5) {};
\node (ZC) at (12, 2.5) {};
\node[draw=none, label = center : {${u}_{2,c}$}] at (11.55, 3.4) {};
\draw[double,->, -{Stealth[length=2mm, width=4mm]}, thick,double distance=2.25pt] (3.2,3) -- (3.8,3);
\path[->, thick] (ZE2) edge[out = -6, in = 160] (ZC);
\node[circle={3pt},draw = black] at (9.5,5.5){$\zeta_1$};
\path[->, thick] (10.2,5.5) edge[] (11.8,5.5);
\path[gray, dashed,->, thick] (10.2,5.5) edge[out = -30, in = 170] (12, 1.5);
\path[gray, dashed,->, thick] (10.2,0.5) edge[] (11.8,-1.5);
\path[->, thick] (10.2,0.5) edge[out = 20, in = -160] (ZC);
\end{tikzpicture}
\caption{Diagram of the extended category model in Description \ref{desc: Main model}}\label{fig: Diagram 2}
\end{figure}

\noindent
Then the ${z}_{i}$'s for $i = 1, \cdots, c+e$ have the same expressions as in \eqref{eq: zeta and z}, and ${\hat{y}}_{k}$, ${\mathcal{L}}_{\textup{CE}}$ are given by the following equations:
\begin{align}\label{eq: hat{y} and L_{CE}}
\hat{y}_{k} = \frac{{e}^{{z}_{k}}}{{e}^{{z}_{1}}+\cdots+{e}^{{z}_{c}}+{e}^{{z}_{c+1}}\cdots+{e}^{{z}_{c+e}}},
{\quad}
{\mathcal{L}}_{\textup{CE}} = 
-({y}_{1}\log{\hat{y}_{1}}+\cdots+{y}_{c}\log{\hat{{y}_{c}}}).
\end{align}
\end{desc}

\vskip 2mm

\begin{remark}\label{remark: approx error}
\rm
Note that the model $\textup{M} \otimes \textup{S}$ can be embedded in the $\textup{M} \otimes (\textup{S} \oplus \textup{S}^{gh})$ for any $e \geq 0$. Hence the approximation errors of $\textup{M} \otimes (\textup{S} \oplus \textup{S}^{gh})$ are less than that of $\textup{M} \otimes \textup{S}$.
\end{remark}

Now, by examining the basic structure of ECM and the first derivative of the loss function, we can derive some properties of ECM. 
From the direct calculations, we obtain the following partial derivatives:
\begin{equation}\label{eq: GD of loss 1}
    \frac{\partial{\mathcal{L}_{\textup{CE}}}}{{\partial{{z}_{i}}}} = 
    \frac{\big(\sum_{j=1, j \neq i}^{c+e} {y}_{j} \big){e}^{{z}_{i}} - {y}_{i}\sum_{j=1, j \neq i}^{c+e} {e}^{{z}_{j}}}{\sum_{j=1}^{c+e} {e}^{{z}_{j}}},
\end{equation}
where $i = 1, \cdots , c$. For the $i = c+1, \cdots, c+e$ cases, the derivatives of the target function is given by the following relations:
\begin{align}\label{eq: GD of loss 2}
    \frac{\partial{\mathcal{L}_{\textup{CE}}}}{{\partial{{z}_{i}}}} = 
    \frac{\big(\sum_{j=1}^{c} {y}_{j} \big){e}^{{z}_{i}}}{\sum_{j=1}^{c+e} {e}^{{z}_{j}}},
\end{align}
where $e \in \mathbb{Z}_{+}$ is the number of the extended categories, called the ghost number. Moreover, from the equation in \eqref{eq: zeta and z}, we have:
\begin{align*}
\frac{\partial{{z}_{i}}}{\partial{\zeta}_{j}} = {u}_{j, i} \cdot \sigma'.
\end{align*}
Note that the derivatives of ${\mathcal{L}_{\textup{CE}}}$ with respect to ${z}_{i}$ for $c+1 \leq i \leq c+e$ also can be thought of as the scalar multiples of the $\hat{y}_{i}$.

\vskip 2mm

Now, consider the situation when $\textup{M} \otimes (\textup{S} \oplus \textup{S}^{gh})$ is sufficiently converges, so that the ${(c + i)}^{\textup{th}}$ values goes to zero, i.e. $\hat{y}_{c+i} \rightarrow {y}_{c+i}=0$ for $ i = 1, \cdots, e$. Equivalently, we know that ${e}^{{z}_{c+i}} \rightarrow 0$, (i.e. ${z}_{c+i} \rightarrow -\infty$). Hence, for the weights ${w}^{k}_{i,j}$ in $\textup{M}$, we obtain the result below:
\begin{align*}
\lVert \frac{{\partial}{\mathcal{L}}_{\textup{M} \otimes (\textup{S} \oplus \textup{S}^{gh})}}{{\partial}{w}^{k}_{i,j}} -
\frac{{\partial}{\mathcal{L}}_{\textup{M} \otimes \textup{S}}}{{\partial}{w}^{k}_{i,j}}
\rVert
&= 
\lVert
\sum_{s=1}^{c}{\frac{{\partial}{\mathcal{L}}_{\textup{M} \otimes (\textup{S} \oplus \textup{S}^{gh})}}{{\partial}{z}_{s}}}{\frac{{\partial}{z}_{s}}{{\partial}{w}^{k}_{i,j}}} 
+ \sum_{s=c+1}^{c+e}{\frac{{\partial}{\mathcal{L}_{\textup{M} \otimes (\textup{S} \oplus \textup{S}^{gh})}}}{{\partial}{z}_{s}}}{\frac{{\partial}{z}_{s}}{{\partial}{w}^{k}_{i,j}}} 
- \sum_{s=1}^{c}{\frac{{\partial}{\mathcal{L}}_{\textup{M} \otimes \textup{S}}}{{\partial}{z}_{s}}}{\frac{{\partial}{z}_{s}}{{\partial}{w}^{k}_{i,j}}}
\rVert
\\
&\leq
\sum_{s=c+1}^{c+e} \lVert
{\frac{{\partial}{\mathcal{L}_{\textup{M} \otimes (\textup{S} \oplus \textup{S}^{gh})}}}{{\partial}{z}_{s}}} \rVert \lVert {\frac{{\partial}{z}_{s}}{{\partial}{w}^{k}_{i,j}}} 
\rVert
+
\sum_{s=1}^{c}
\lVert
{\frac{{\partial}{\mathcal{L}}_{\textup{M} \otimes (\textup{S} \oplus \textup{S}^{gh})}}{{\partial}{z}_{s}}} - {\frac{{\partial}{\mathcal{L}}_{\textup{M} \otimes \textup{S}}}{{\partial}{z}_{s}}} \rVert \lVert {\frac{{\partial}{z}_{s}}{{\partial}{w}^{k}_{i,j}}}
\rVert
\rightarrow 0,
\end{align*}
from the equations in \eqref{eq: GD of loss 1} and \eqref{eq: GD of loss 2}.

\begin{result}\label{theorem: GD}
\rm
Let $\textup{M} \otimes (\textup{S} \oplus \textup{S}^{gh})$ be an ECM given in Description \ref{desc: Main model}. Suppose that $\lVert {\frac{{\partial}{z}_{s}}{{\partial}{w}^{k}_{i,j}}}
\rVert$ is finite for $s = 1, ... , c+e$. Then, after the model $\textup{M} \otimes (\textup{S} \oplus \textup{S}^{gh})$ converges sufficiently, the gradients become the same as those of the original model $\textup{M} \otimes \textup{S}$.
\end{result}

\vskip 2mm

As seen in Remark \ref{remark: approx error} and Result \ref{theorem: GD} above, ECM is an extension of the original model, and after sufficient convergence, it is expected to exhibit similar behavior. However, $\textup{M} \otimes (\textup{S} \oplus \textup{S}^{gh})$ has the degree of freedom for various descent directions in the early stages of training. Moreover:
\begin{result}
\rm
The gradient of $\textup{M} \otimes (\textup{S} \oplus \textup{S}^{gh})$ is differ from the gradient of $\textup{M} \otimes \textup{S}$ in the early stages of training.
\end{result}

\noindent
Indeed, from \eqref{eq: GD of loss 1} and \eqref{eq: GD of loss 2}, the derivative of the loss function of $\textup{M} \otimes (\textup{S} \oplus \textup{S}^{gh})$ is given by the following relation:
\begin{align*}
\frac{{\partial}{\mathcal{L}}_{\textup{M} \otimes (\textup{S} \oplus \textup{S}^{gh})}}{{\partial}{w}^{k}_{i,j}}
&= \sum_{s=1}^{c}\frac{\big(\sum_{j=1, j \neq s}^{c} {y}_{j} \big){e}^{{z}_{s}} - {y}_{s}\big(\sum_{j=1, j \neq s}^{c} {e}^{{z}_{j}}\big) - {y}_{s}\big(\sum_{j= c+1}^{c+e} {e}^{{z}_{j}}\big)}{\big(\sum_{j=1}^{c} {e}^{{z}_{j}}\big) + \big(\sum_{j=c+1}^{c+e} {e}^{{z}_{j}}\big)}{\frac{{\partial}{z}_{s}}{{\partial}{w}^{k}_{i,j}}}
\\
&+ \sum_{s=c+1}^{c+e}\frac{\big(\sum_{j=1}^{c} {y}_{j} \big){e}^{{z}_{s}}}{\big(\sum_{j=1}^{c} {e}^{{z}_{j}}\big) + \big(\sum_{j=c+1}^{c+e} {e}^{{z}_{j}}\big)}{\frac{{\partial}{z}_{s}}{{\partial}{w}^{k}_{i,j}}},
\end{align*}
and, the derivative of the loss function for the original model $\textup{M} \otimes \textup{S}$ is:
\begin{align*}
\frac{{\partial}{\mathcal{L}}_{\textup{M} \otimes \textup{S}}}{{\partial}{w}^{k}_{i,j}}
&= \sum_{s=1}^{c}\frac{\big(\sum_{j=1, j \neq s}^{c} {y}_{j} \big){e}^{{z}_{s}} - {y}_{s}\big(\sum_{j=1, j \neq s}^{c} {e}^{{z}_{j}}\big)}{\sum_{j=1}^{c} {e}^{{z}_{j}}}{\frac{{\partial}{z}_{s}}{{\partial}{w}^{k}_{i,j}}}.
\end{align*}
Hence the direction of descent for $\textup{M} \otimes (\textup{S} \oplus \textup{S}^{gh})$ is indeed differ from that of the original model $\textup{M} \otimes \textup{S}$.

\vskip 2mm

Let $w\in W\subset\mathbb R^{d}$ collect all original weights and set $\gamma \in B$ for the connections that feed the network output to the ghosts (i.e., the ghost parameter $\gamma$ is the notation that group the parameters ${u}_{j, c+i}$ in Figure \ref{fig: Diagram 2}). The full parameter is $(w,\gamma)\in\widetilde W:=W \times B$, where a bounded space $B \subset {\mathbb{R}}^{e \times s}$ keeps $\widetilde W$ compact. For a training sample $(x,y)$ with label $y\in\{1,\dots,c\}$ denote the outputs by $z_{1},\dots,z_{c+e}$ as in \eqref{eq: zeta and z}. Define the per-sample losses
\begin{align*}
  \mathcal{L}_{\textup{ext}}\bigl((w,\gamma);x,y\bigr) = -\log
       \frac{e^{z_{y}}}{\big(\sum_{j=1}^{c} {e}^{{z}_{j}}\big) + \big(\sum_{j=c+1}^{c+e} {e}^{{z}_{j}}\big)}, \quad
  \mathcal{L}_{\textup{orig}}\bigl(w;x,y\bigr) = -\log
       \frac{e^{z_{y}}}{\sum_{j=1}^{c} {e}^{{z}_{j}}}.
\end{align*}
Then, their difference is the ghost contribution term:
\begin{align*}
\mathcal{L}_{\textup{ghost}}\bigl( (w, \gamma) ; x \bigr) = \log\,\Bigl(1 + \frac{\big(\sum_{j=c+1}^{c+e} {e}^{{z}_{j}}\big)}{\sum_{j=1}^{c} {e}^{{z}_{j}}}\Bigr)\,\ge 0,
\end{align*}
which is independent of the class label $y$. Averaging over the data set yields the objective function $f_{\textup{ext}} = f_{\textup{orig}} + f_{\textup{ghost}}$ on $\widetilde W$. With a fixed step size $\eta > 0$, let $\widetilde g$ be a mini-batch estimator of $\nabla f_{\textup{ext}}$. Then the update rule for the SGD map
\begin{align*}
({w}^{(t+1)},{\gamma}^{(t+1)}) = ({w}^{(t)},{\gamma}^{(t)}) - \eta\,\widetilde g\bigl(({w}^{(t)},{\gamma}^{(t)}),\xi_t\bigr), \qquad t \ge 0,
\end{align*}
defines a Markov chain on $\widetilde W$.

As in the non-ghost case, the continuity of the update map and compactness of $\widetilde W$ imply the existence of at least one invariant probability measure $\widetilde\mu$ (see Section 3).
For any bounded continuous $h:\widetilde W\to\mathbb R$ we have
\[
  \lim_{N\to\infty}\frac1N\sum_{t=0}^{N-1}h({w}^{(t)},{\gamma}^{(t)})
  =\int_{\widetilde W}h\,d\widetilde\mu
  \quad\text{almost surely}.
\]
Choosing $h=f_{\textup{orig}}$ and $h=f_{\textup{ghost}}$ gives
\begin{align}\label{erg : ghost}
  \frac1N\sum_{t=0}^{N-1}f_{\textup{orig}}({w}^{(t)})
  \longrightarrow \int f_{\textup{orig}}\,d\widetilde\mu,
  \qquad
  \frac1N\sum_{t=0}^{N-1}f_{\textup{ghost}}({w}^{(t)},{\gamma}^{(t)})
  \longrightarrow \int f_{\textup{ghost}}\,d\widetilde\mu.
\end{align}

At the start of training $z_{c+i}\approx 0$, so
$f_{\textup{ghost}}$ is comparable to $f_{\textup{orig}}$ and the
gradients $\partial f_{\textup{ghost}} / \partial \gamma_i$
add extra descent directions that help the optimizer escape poor basins. As optimization proceeds, $\hat y_{c+i}$ decay to zero, causing $\mathcal{L}_{\textup{ghost}}$ to vanish.  
With bounded Jacobians and the gradient formulas in
\eqref{eq: GD of loss 1}–\eqref{eq: GD of loss 2}, the gradients of the
extended and original models coincide up to negligible terms.  
Hence
\begin{align}\label{erg: ghost2}
  0 \;\le\; \int f_{\textup{ghost}}\,d\widetilde\mu
  \;\le\; C(M,e),
\end{align}
where $C(M,e) \rightarrow 0$, so that the ghost component has no adverse long-term effect while providing useful freedom of exploration during the initial training phase.

However, during the initial phase of training, there is an almost constant negative gradient in the $\gamma$ direction.
Even if the original loss $f_{\mathrm{orig}}$ is surrounded inside
$W$ by a saddle ridge of height $\varepsilon$, $f_{\mathrm{orig}}$ itself does not change along the $\gamma$ axis, $f_{\mathrm{ghost}}$ decreases along that axis. Hence, the level set of constant total loss in the extended parameter space opens sideways. Therefore, the extra dimensions provide a path that bypasses the barrier $\varepsilon$ inside $W$. A slightly more mathematical restatement of the above contents is as follows:

\begin{result}\rm\label{curve}
There exists a piecewise $C^{1}$ curve
\[
  (w(t),\gamma(t))_{t\in[0,1]}\subset\widetilde W
  \]
with $\frac{d}{dt}\,f_{\mathrm{ext}}\bigl(w(t),\gamma(t)\bigr)\le 0$,
such that
\[
  f_{\mathrm{orig}}\bigl(w(1)\bigr)
      < f_{\mathrm{orig}}\bigl(w(0)\bigr)-\varepsilon .
\]
\end{result}

Now, we present the experimental result of a small toy example designed to demonstrate that adding ghost nodes can actually shortcut the training trajectory and reach a good solution in fewer epochs. For MNIST, we drew 30 training samples per class (300 total, that is, RMNIST/30 data) without replacement in each run and added two ghost nodes. We repeated this experiment 30 times. Below is the mean test loss $\pm1\sigma$ band plot. On average, the epoch at which the first local maximum peak (see, for example, \citet{Belkin19} and \citet{Nakkiran20}) occurs was 4.8 epochs earlier with ghost nodes than without. Note that the effect observed below may diminish with sufficiently large datasets. However, since such phenomena can still occur, more rigorous experiments are necessary. See Appendix \ref{sec:app A} for the individual results of each RMNIST/30 experiment. In this toy example, training is conducted using pure gradient descent (full-batch, to clearly observe different trajectories) with zero momentum. We adopt as the classifier a lightweight CNN consisting of two convolution–max-pooling blocks, each block using a $3\times3$ kernel and a pooling size of $2\times2$.

\begin{figure}[htbp]
  \centering
  \begin{subfigure}[b]{0.48\textwidth}
    \centering
    \includegraphics[width=\textwidth]{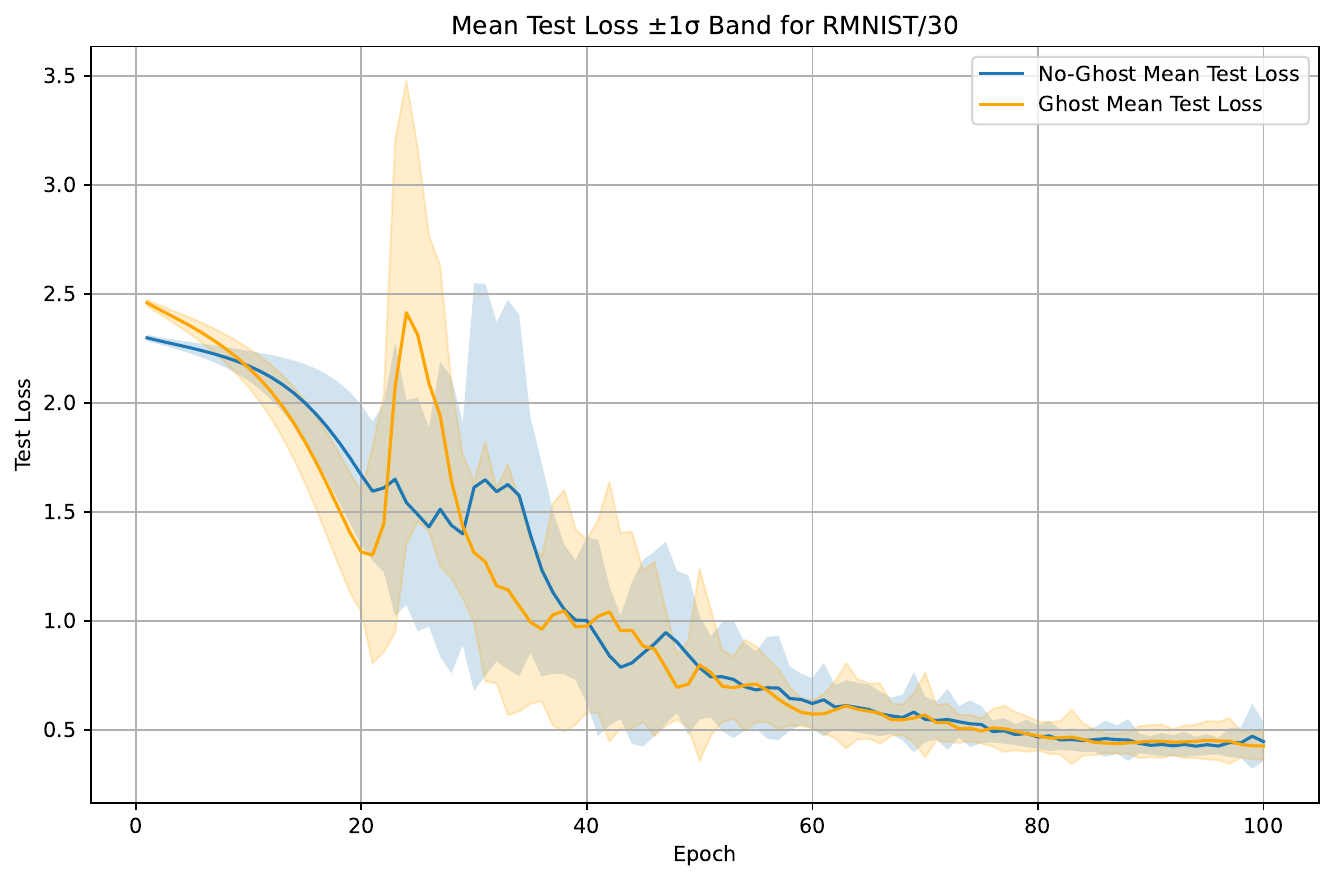}
    \caption{Mean Test Loss $\pm1\sigma$ Band for RMNIST/30}
  \end{subfigure}
  \hfill
  \begin{subfigure}[b]{0.48\textwidth}
    \centering
    \includegraphics[width=\textwidth]{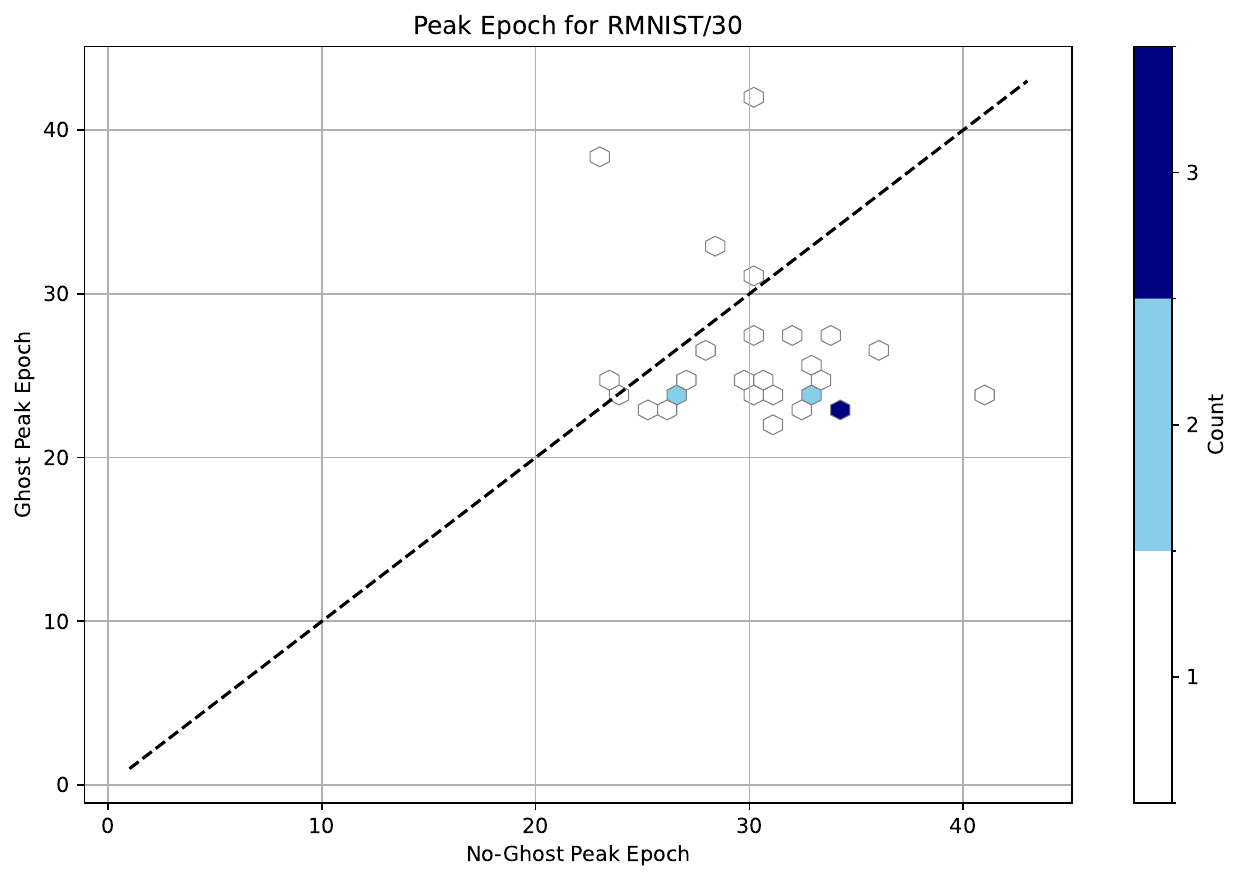}
    \caption{Peak Epoch Comparison for RMNIST/30}
  \end{subfigure}
  \caption{%
    (a) Mean test loss curves with $\pm1\sigma$ shading 
    (b) Peak epoch comparison%
  }
  \label{fig:rmnist30}
\end{figure}\label{fig: RMNIST30}

To summarize (see equation \eqref{erg : ghost}, \eqref{erg: ghost2} and Result \ref{curve}), the key idea is to bypass barriers early on and quickly move toward a local minimizer via a gentle shortcut. In the later stages, as ghost nodes are added, the expanded parameter space contracts along the ghost axis, and the resulting function becomes almost identical to the original one (up to minor noise differences). According to the application of the ergodic theorem discussed above, after many training steps, the average loss eventually converges to a similar value. Ultimately, the core of the ghost extension is to guide the optimization toward the original convergence point in advance with fewer steps. Thus, although the $w$ component alone does not cross the barrier, the curve uses the $\gamma$ coordinates to reach a region where the original loss is lower.

In other words, Result \ref{curve} shows that the added ghost coordinates open a lateral corridor in the enlarged parameter space, allowing the SGD trajectory to bypass small ridges that would otherwise trap it in the original weight domain $W$. From an ergodic perspective, this enlargement may increase its spectral gap, so that it accelerates mixing toward the invariant distribution. In the early stages, the marginal distribution over the original weights might resemble a smoothed convolution, since the chain first disperses mass along the ghost directions before projecting back to $W$. As optimization proceeds, the contribution allocated to the ghost dimensions vanishes, and it is proven that the marginal converges to the same stationary measure governing the unextended model. Such behavior is also expected to be reflected in numerical diagnostics such as the Lyapunov exponent, as the ghost coordinates accelerate escape from shallow basins; consequently, the Lyapunov exponent could turn negative earlier, indicating an earlier entry into a contractive state.

\vskip 4mm

\section{Conclusion}\label{sec:contribution and future}

The fact that the entire SGD trajectory admits an invariant probability measure whose ergodic averages capture both the loss landscape and the gradient geometry of deep networks has been extensively studied in a series of recent works. Based on these studies, we propose that the Lyapunov-based indicator $\hat\gamma_N$ provides a practical test for distinguishing true convergence from stagnation at saddle points.

As a new approach to improving training efficiency, we propose attaching ghost nodes that temporarily expand the output space to guide optimization—creating escape directions that accelerate early convergence—while smoothly vanishing over time and acting as a built-in regularizer; this mechanism extends naturally to ghost-augmented classifiers, explaining their empirically observed boost in early-stage trainability.

Looking forward, the ghost formalism itself suggests several promising directions.  Because ghost dimensions effectively flatten narrow barriers, one may exploit them to perform finer-grained, ergodic sampling of weight space, enabling more detailed statistical probes of generalized loss surfaces than are currently possible.  In particular, we expect that ghost-assisted ergodic diagnostics could reveal how often SGD visits regions that are invisible in the original parameterization and thereby refine our understanding of implicit regularization.

Several avenues for future work naturally emerge. A first priority is to conduct large-scale empirical studies to test whether the same guarantees persist under mini-batch SGD and adaptive optimization schemes (see \citet{ZLSJ22}, \citet{Zhang23}, \citet{VGCX24}). In parallel, a detailed ergodic analysis of the saddle-point landscape together with investigations into the effectiveness of noise injection, momentum, and related acceleration techniques could deepen our understanding of optimization dynamics in modern networks. Furthermore, by integrating both the parameter-space dynamics for specific objective functions and the learning rate into a single unified framework, one can undertake more detailed analyses using advanced ergodic-theoretic concepts rather than relying solely on its basic formalism.

Moreover, focusing on our main result, we expect to further extend the ghost extension to a wider variety of application domains. Indeed, it is important to extensively evaluate the stability and performance improvements provided by ghost extensions across diverse settings, such as image classification benchmarks (CIFAR, ImageNet), NLP sequence tagging tasks, time series forecasting, graph neural networks, and multimodal learning.

Additionally, there are opportunities to simplify hyperparameter tuning even further. By automating or reducing dependency on additional hyperparameters such as momentum or batch size—not just relying on a fixed learning rate combined with ghost structures—using methods such as one-shot tuning or Bayesian optimization, we can significantly lower the barrier to applying ghost extensions to practical engineering problems.

Next, it is promising to explore how ghost nodes interact with existing regularization and optimization techniques. By combining ghost extensions with Label Smoothing, BatchNorm variants, Mixup, Dropout, or adaptive optimizers, we may uncover synergistic effects that further enhance generalization performance and robustness against overfitting. Furthermore, we are interested in designing adaptive ghost strategies that dynamically adjust the number or scale of ghost dimensions throughout training. Allocating more ghost capacity during initial exploratory stages and gradually reducing it as the model approaches convergence could lead to a more efficient learning trajectory.

\newpage
\appendix
\FloatBarrier

\section{Individual Experimental Results for RMNIST/30}\label{sec:app A}

The figures below are the results of experiments in which we formed 300 training subsets from the MNIST training set by drawing 30 samples per class without replacement. We ran a baseline model with no ghost nodes and a model augmented with two ghost nodes. In order, are plots of the training loss, the test loss, and finally the test accuracy over 30 independent runs.

\begin{figure}[h]
  \centering
      \includegraphics[
    width=16.5cm,
    height=18cm
  ]{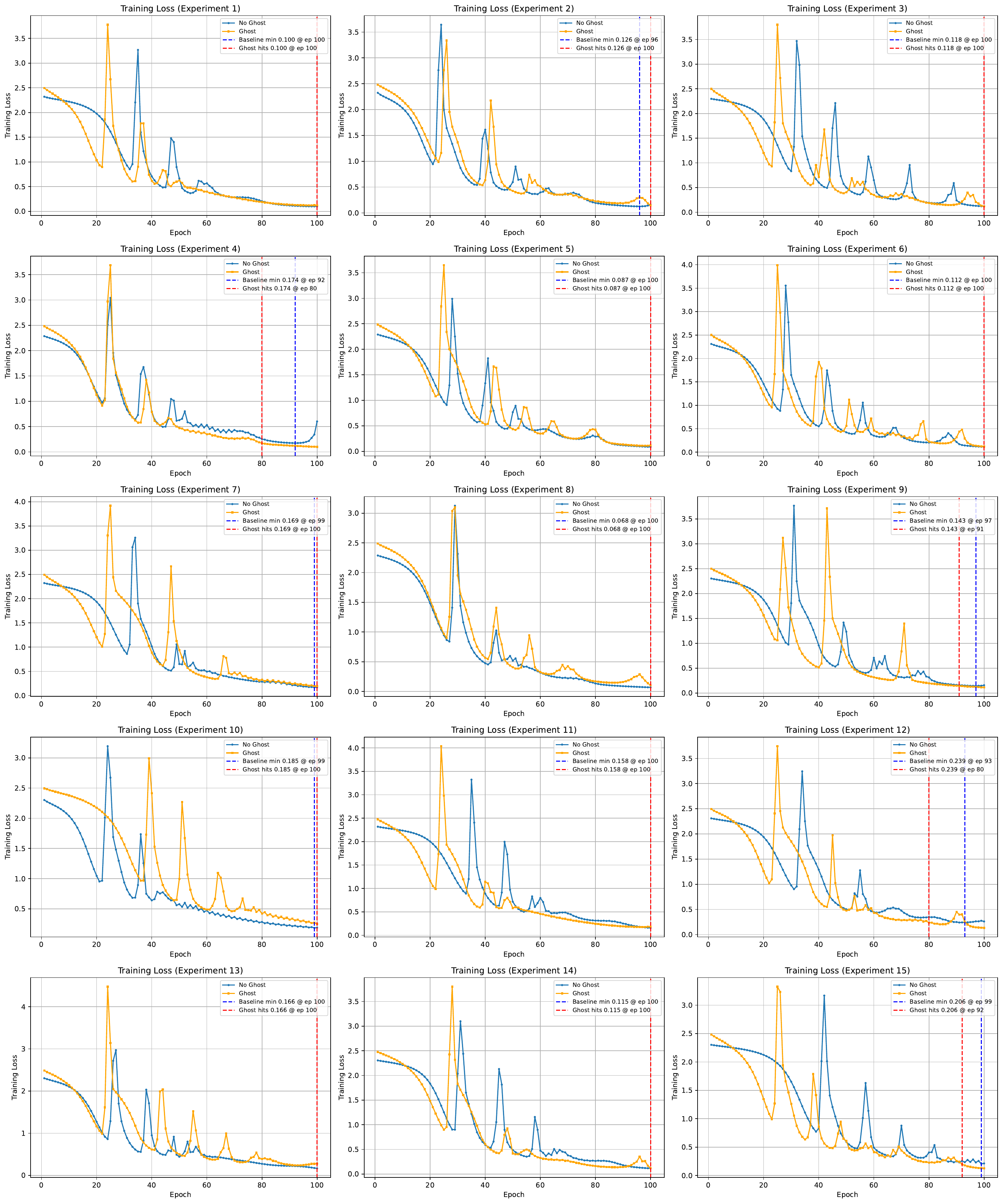}
\caption{Training Loss for Experiment 1 to Experiment 15}
\end{figure}

\begin{figure}[htbp]
  \centering
    \includegraphics[width=\textwidth]{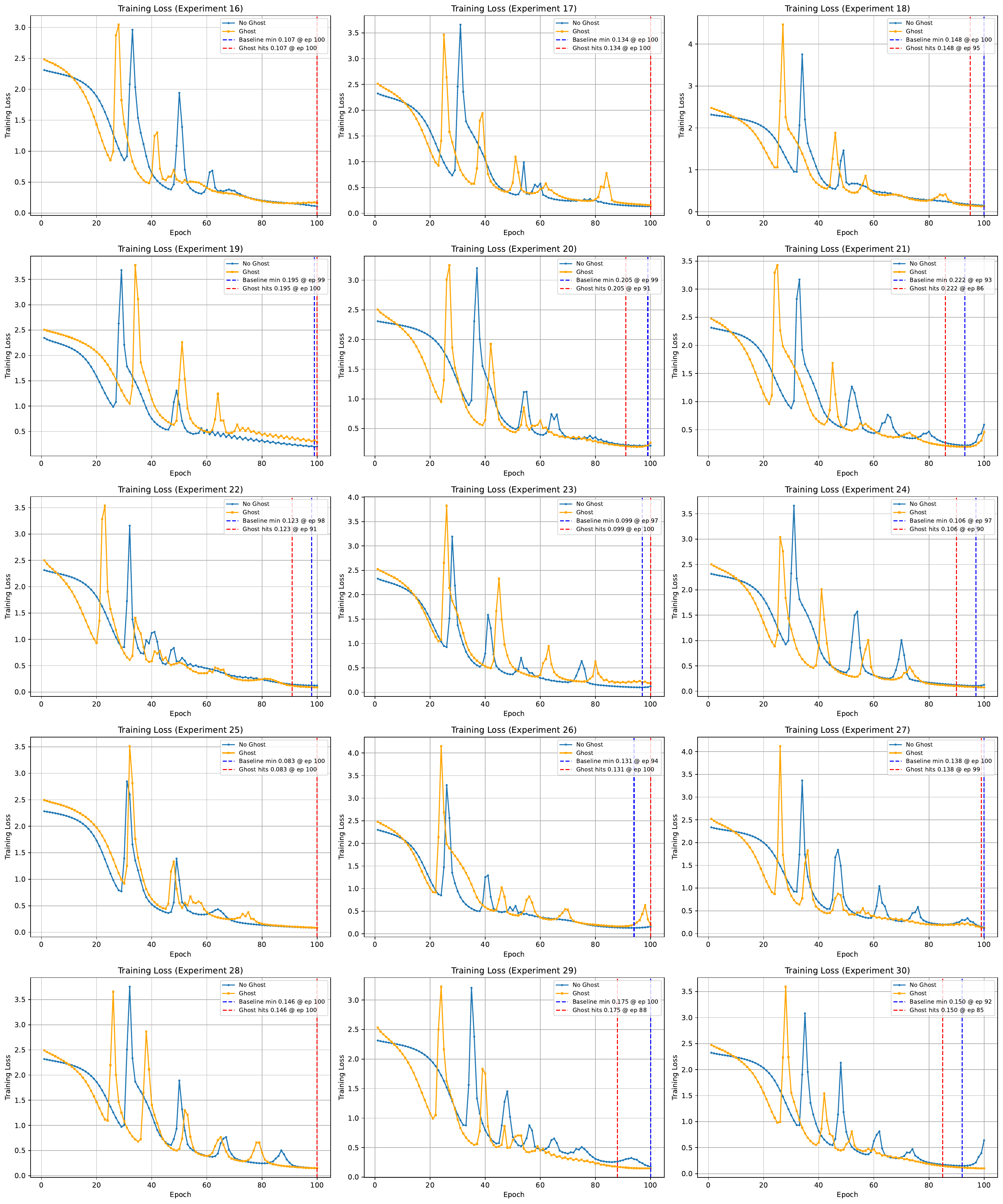}
\caption{Training Loss for Experiment 16 to Experiment 30}
\end{figure}

\begin{figure}[htbp]
  \centering
    \includegraphics[width=\textwidth]{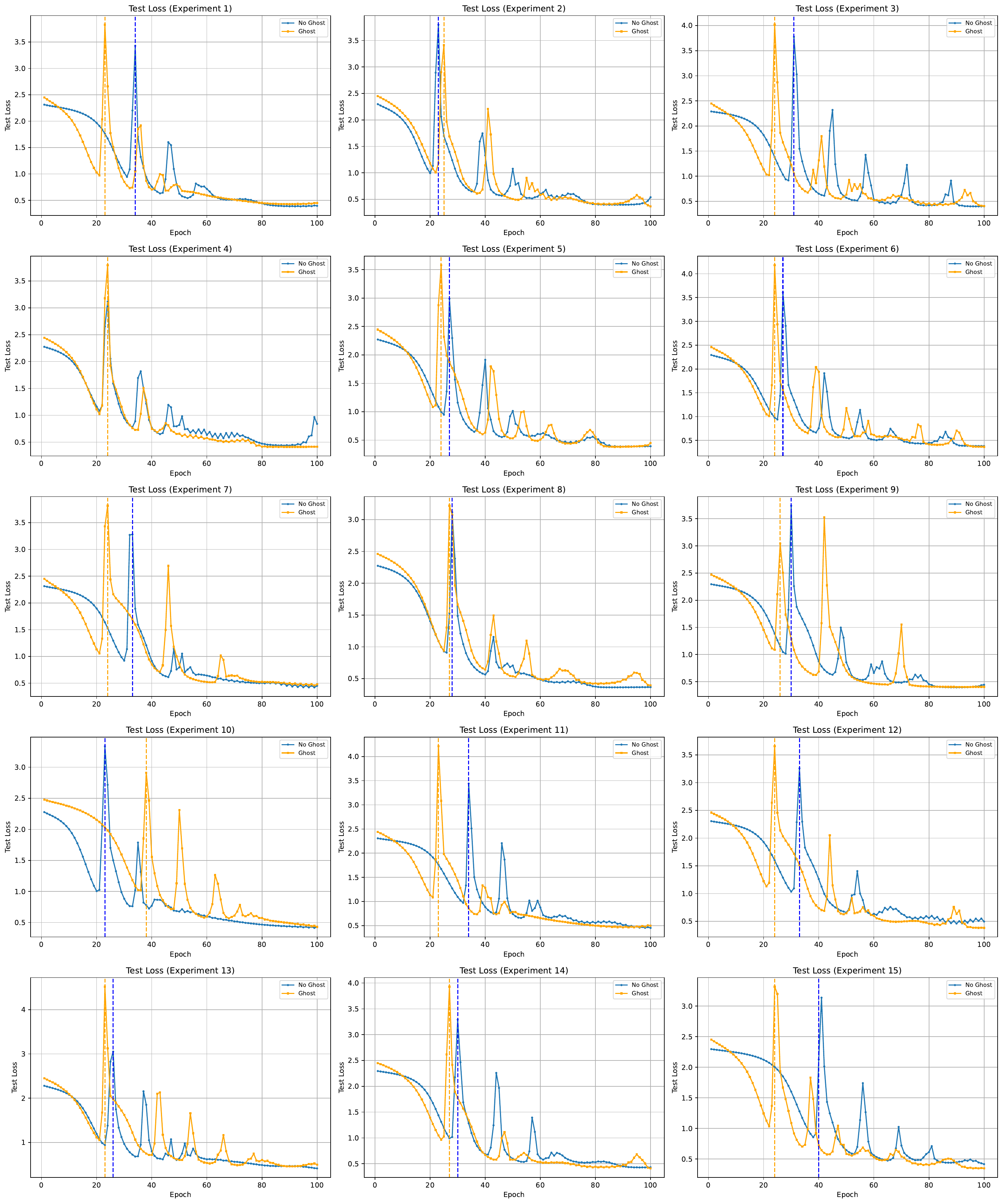}
\caption{Test Loss for Experiment 1 to Experiment 15}
\end{figure}

\begin{figure}[htbp]
  \centering
    \includegraphics[width=\textwidth]{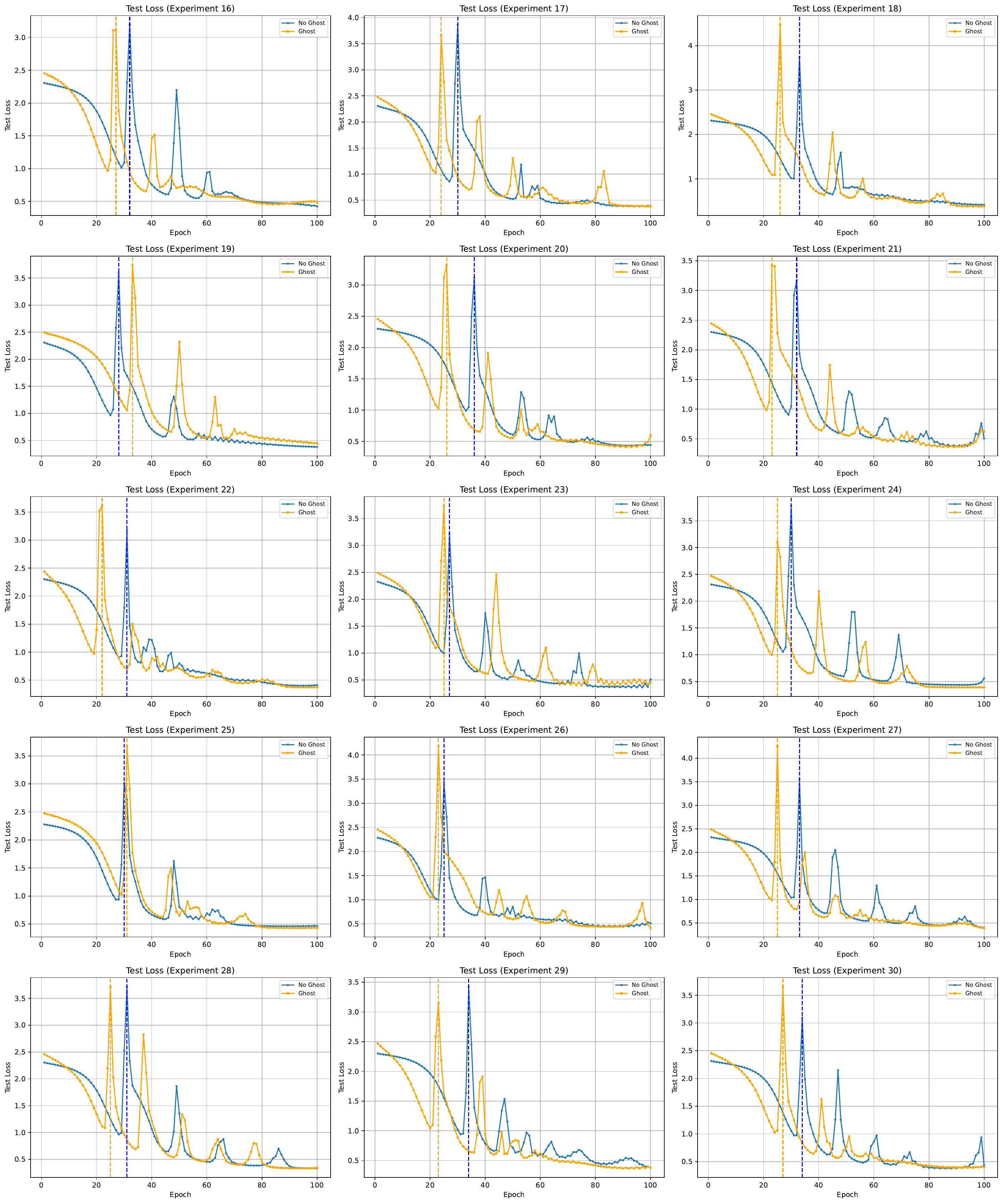}
\caption{Test Loss for Experiment 16 to Experiment 30}
\end{figure}

\begin{figure}[htbp]
  \centering
    \includegraphics[width=\textwidth]{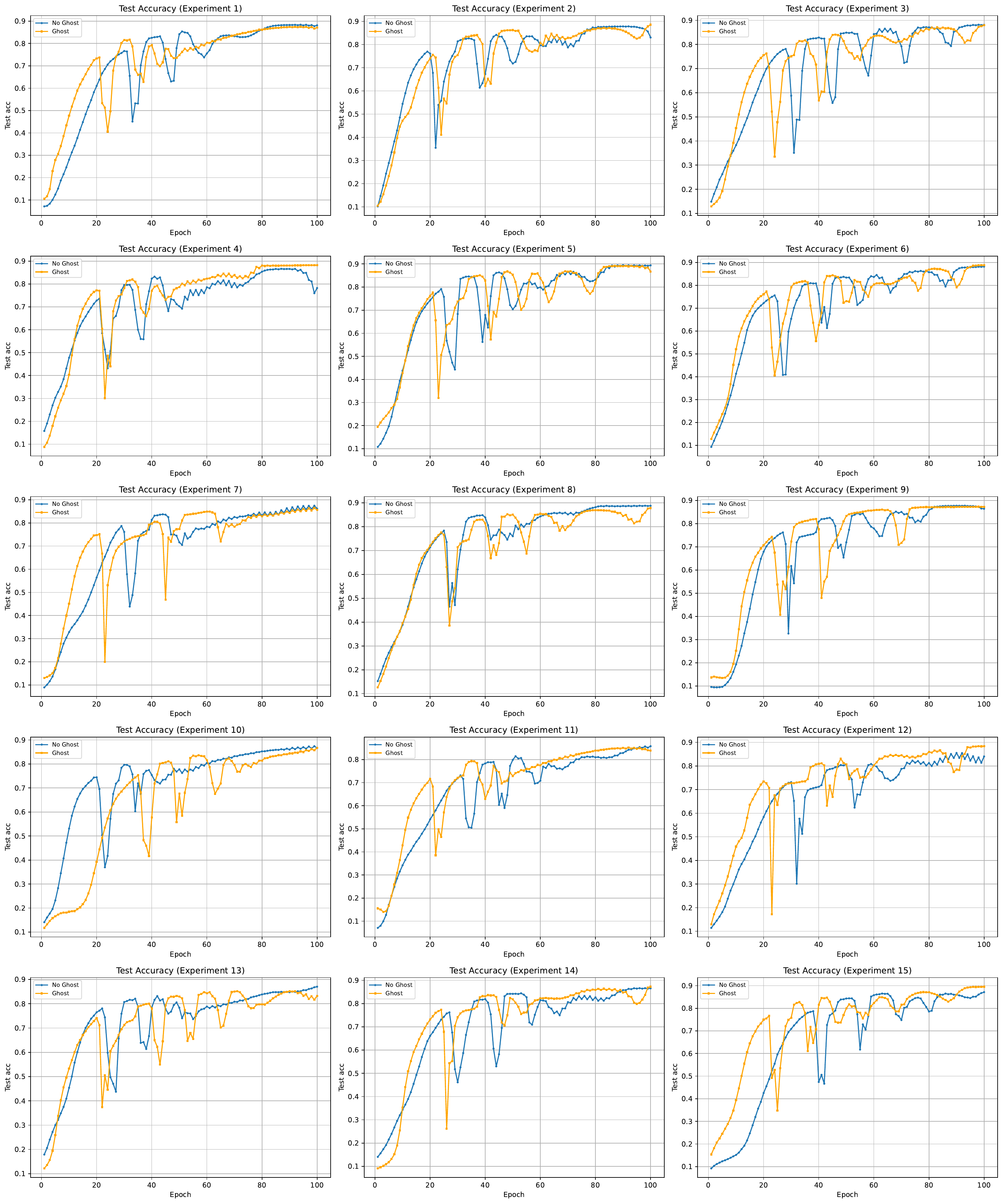}
\caption{Test Accuracy for Experiment 1 to Experiment 15}
\end{figure}

\begin{figure}[htbp]
  \centering
    \includegraphics[width=\textwidth]{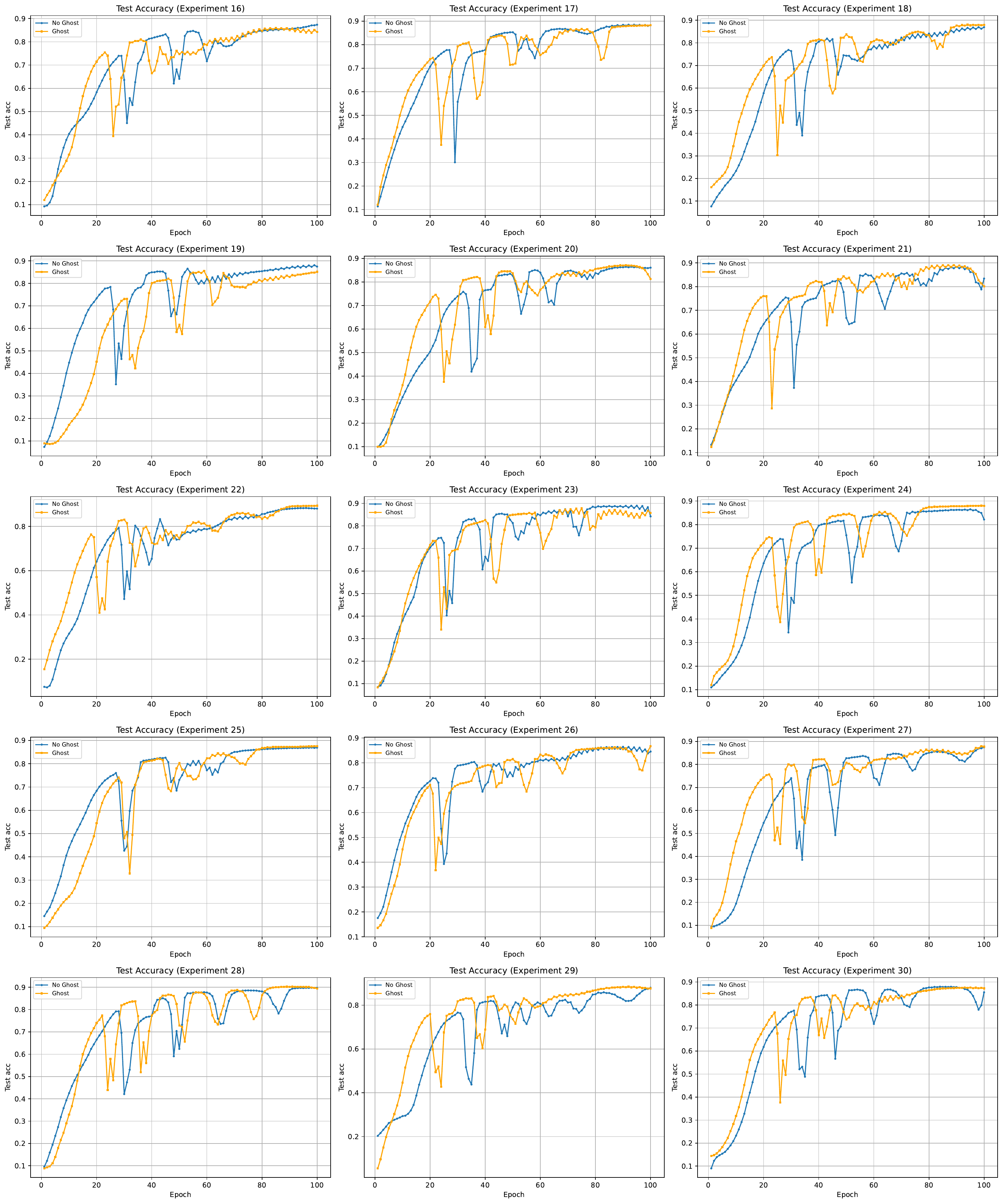}
\caption{Test Accuracy for Experiment 16 to Experiment 30}
\end{figure}

\end{document}